\documentclass[twoside,leqno,twocolumn]{article}

\usepackage[letterpaper]{geometry}

\usepackage{ltexpprt}
\usepackage{hyperref}
\usepackage{graphicx}

\begin{document}

\newcommand\relatedversion{}

\title{\Large Making a Computational Attorney}
\author{Dell Zhang\thanks{Thomson Reuters Labs} \thanks{\texttt{dell.z@ieee.org}, \texttt{frank.schilder@thomsonreuters.com}} \and Frank Schilder$^{*\dag}$ \and Jack G. Conrad$^*$ \and Masoud Makrehchi$^*$ \and David von Rickenbach$^*$ \and Isabelle Moulinier$^*$}


\date{}
\maketitle





\fancyfoot[R]{\scriptsize{Copyright \textcopyright\ 2023\\
Copyright retained by principal author's organization}}


\begin{abstract} \small\baselineskip=9pt 
This ``blue sky idea'' paper outlines the opportunities and challenges in data mining and machine learning involving making a \emph{computational attorney} --- an intelligent software agent capable of helping human lawyers with a wide range of complex high-level legal tasks such as drafting legal briefs for the prosecution or defense in court. 
In particular, we discuss what a ChatGPT-like Large Legal Language Model (L$^3$M) can and cannot do today, which will inspire researchers with promising short-term and long-term research objectives.
\end{abstract}

\section{Introduction}
\label{sec:Introduction}

The legal domain has always been an important application area of cutting-edge \emph{data mining} and \emph{machine learning} techniques. 
For the last three decades, research on legal AI has also been pushing the frontier of data mining and machine learning~\cite{governatoriThirtyYearsArtificial2022,sartorThirtyYearsArtificial2022,villataThirtyYearsArtificial2022,bench-caponHistoryAILaw2012,surdenMachineLearningLaw2014}. 
In comparison with other application areas, work with legal data is characterized by the following unique features.
\begin{itemize}
  \item \emph{Massive scale of complex text data.} 
  For example, Thomson Reuters has accumulated over 60,000 TBs worth of data, a substantial portion of which is legal text data. 
  There are also publicly available large legal corpora such as the 256GB Pile-of-Law dataset~\cite{hendersonPileLawLearning2022}. 
  The majority of such data are in the form of long documents written in formal and professional language, such as legal judgments, legal opinions, and legal contracts. 
  \item \emph{High labeling cost.}
  For example, attorney fees in the US usually range from about \$100 to \$1,000 or more per hour, which makes the acquisition of gold-standard annotations from legal \emph{subject matter experts} (SMEs) for training and testing legal AI models very expensive.
  \item \emph{Emphasis on thoroughness as well as precision.}
  For example, legal research platforms like Westlaw and e-discovery tools usually put a very high weight on \emph{recall} while also demanding \emph{precision} (see the latest Westlaw Precision release).
  \item \emph{Requirement of specialist knowledge.}
  For example, to qualify as an attorney in the US, one must typically complete 7+ years of post-secondary education including 3 years at an accredited law school and then pass a difficult professional-license exam commonly known as the ``bar exam''~\cite{bommaritoiiGPTTakesBar2022}.
\end{itemize}

Apparently, the rise of \emph{pre-trained large language models} such as BERT and GPT is causing a \emph{paradigm shift} in data mining and machine learning.
The legal domain is no exception. 
Hence, we propose to reexamine the current research agenda on legal AI by rethinking the question ``\underline{what does it take to make a \emph{computational attorney}?}'' 
The reason we coin a new term ``computational attorney'' instead of using the existing term ``computational law'' is to stress our anticipation for such capabilities to go beyond the automation of legal compliance management (e.g., based on \emph{computable contracts}~\cite{surdenComputableContracts2012}) or other mundane legal information processing tasks (typically carried out by \emph{paralegals}) and help human lawyers with complex high-level legal tasks (like drafting legal briefs for the prosecution or defense in court). 
Such computational attorneys are expected not to replace human lawyers but to work as their competent and reliable partners\footnote{\url{https://tinyurl.com/y9zggu6n}}.

The success of this vision will change the legal industry (of \$300+ billion annual revenue in the US) in at least two aspects: 
(i)~drastically improving the efficiency of millions of attorneys as well as law firms; and 
(ii)~democratizing the legal services~\cite{bexIntroductionSpecialIssue2017} in this law-dependent world where as many as 86\% of low-income Americans with civil legal problems report inadequate or no legal assistance due to prohibitively expensive legal fees~\cite{bommasanietalOpportunitiesRisksFoundation2022}.
Conversely, the opportunities and challenges posed by the pursuit of a computational attorney can also motivate basic research objectives for data mining and machine learning which we will sketch out below.

\section{Legal AI Research: The Trajectory}
\label{sec:Trajectory}

\subsection{Past}
\label{sec:Past}

The usage of data mining and machine learning for legal tasks such as computer-assisted classification of legal abstracts dates back to 1990s~\cite{yang-stephensComputerAssistedClassificationLegal1999}.
In the past 30 years~\cite{governatoriThirtyYearsArtificial2022,sartorThirtyYearsArtificial2022,villataThirtyYearsArtificial2022,bench-caponHistoryAILaw2012,surdenMachineLearningLaw2014}, a variety of data mining and machine learning techniques have been applied to automate relatively minor and repetitive low-level legal tasks, e.g., 
legal text classification and summarization, 
information extraction from legal documents, 
similar case matching, 
and litigation analytics. 
Major data mining conferences such as ICDM have recently included workshops dedicated to this research field\footnote{\url{https://www.mlld.cc/}}.  
Usually, a specific model will be developed to address a specific legal task.

\subsection{Present}
\label{sec:Present}

The research on legal AI is undergoing a fundamental change at the moment: 
instead of \emph{many small models} each for one specific task, researchers have started to build and utilize \emph{one big model} for many different tasks.
Such a big model, aka \emph{foundation model}~\cite{bommasanietalOpportunitiesRisksFoundation2022}, for the legal domain, is a large language model either pre-trained on legal corpora from scratch or adapted from a general model with further pre-training on legal corpora~\cite{chalkidisLEGALBERTMuppetsStraight2020,zhengWhenDoesPretraining2021}, which we call a Large Legal Language Model (L$^3$M). 

The characteristics of the legal domain mentioned in Section~\ref{sec:Introduction} suggest that L$^3$M has huge advantages over traditional technical approaches to legal AI problems.
On one hand, the massive scale of complex text data enables or facilitates the (self-supervised) pre-training of L$^3$M.
On the other hand, the \emph{few-shot} prompting (i.e., \emph{in-context learning}) or \emph{zero-shot} prompting capability of L$^3$M for downstream tasks can greatly alleviate or even avoid the high labeling cost, while the flexibility of L$^3$M to accommodate ambiguity and idiosyncrasies can help to meet the challenges of thoroughness and specialized knowledge.
It is not surprising that with L$^3$Ms such as LEGAL-BERT~\cite{chalkidisLEGALBERTMuppetsStraight2020} and Lawformer~\cite{xiaoLawformerPretrainedLanguage2021}, we are seeing new heights achieved in legal text classification and other tasks~\cite{zhengWhenDoesPretraining2021,songEffectivenessPreTrainedLanguage2022}.

\begin{figure}[!tb]
  \includegraphics[width=\columnwidth]{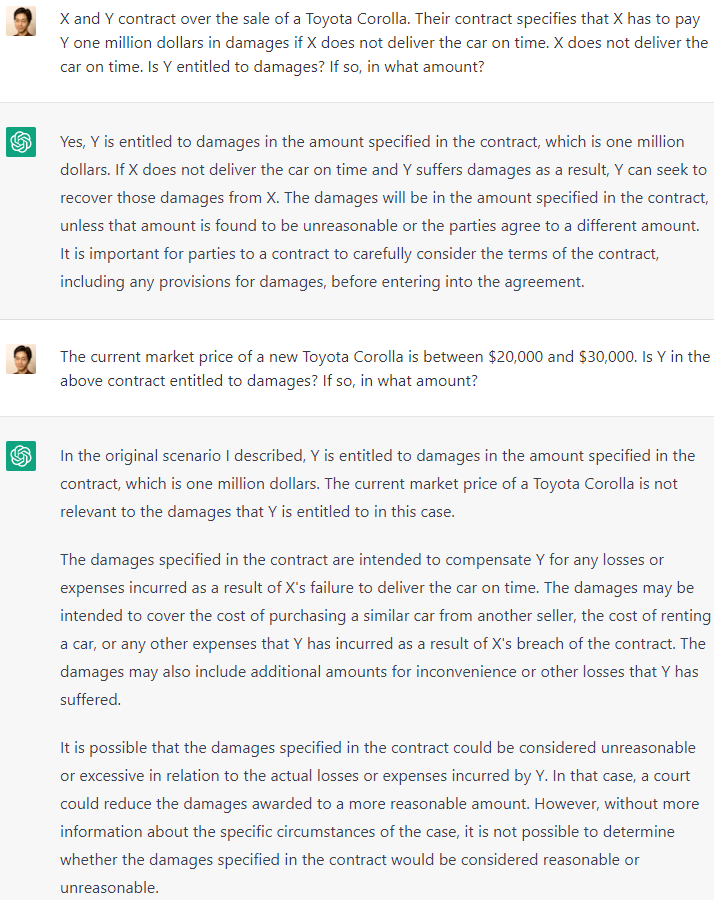}
  \caption{ChatGPT exhibits the sign of preliminary \emph{legal reasoning} in answering a question about ``liquidated damages'' after additional information is supplied, which is beyond the ability of GPT-3 shown in~\cite{bommasanietalOpportunitiesRisksFoundation2022}. }
  \label{fig:chatgpt}
\end{figure}

More importantly, when the scale of a L$^3$M goes above a certain phase-change threshold, it will start to show some \emph{emergent abilities}~\cite{weiEmergentAbilitiesLarge2022} that are present in only very large models but not smaller ones. 
It seems that \emph{legal reasoning} (which is largely beyond the reach of traditional techniques) is one of these emergent abilities. 
L$^3$M based \emph{legal prompt engineering} (aka \emph{legal prompting})~\cite{yuLegalPromptingTeaching2022,trautmannLegalPromptEngineering2022} has demonstrated impressive performance and great potential on COLIEE entailment tasks where the questions come from Japan's rigorous bar exam (which has less than a 40\% passing rate in 2020). 
Furthermore, OpenAI's recently released GPT-3.5 and ChatGPT\footnote{\url{https://openai.com/blog/chatgpt/}}, though not adapted to the legal domain, has shown the preliminary ability of \emph{legal reasoning} which is lacking in its previous version GPT-3 (see Figure~\ref{fig:chatgpt}). 
A comprehensive study of GPT-3.5 (\texttt{text-davinci-003}) shows that it achieved an average accuracy of above 50\% on the complete MBE section of the National Conference of Bar
Examiners (NCBE) practice exams~\cite{bommaritoiiGPTTakesBar2022}, significantly exceeding the 25\% baseline guessing rate. 
It has also been reported that ChatGPT could have passed a practice US bar exam with a headline score of 70\% (35/50)\footnote{\url{https://tinyurl.com/2m6yewed}}.
If such general-purpose models are further enhanced by incorporating legal domain knowledge~\cite{zhenSurveyKnowledgeEnhancedPretrained2022}, they will be even more proficient at legal tasks.

Therefore, we now have good reason to believe that it would not take too long for a ChatGPT style L$^3$M to carry out the above-mentioned simple low-level legal tasks (see Section~\ref{sec:Past}) with human-like or superhuman performance, and thus render the corresponding problems trivial for scientific research. 

So, what's next?

\subsection{Future}
\label{sec:Future}

Although L$^3$Ms have come a long way, we think that there is still a steep path ahead of them to master the intricate art of legal reasoning.
Other than the commonly desired properties like \emph{fast}, \emph{economical}, \emph{accurate}, \emph{interpretable}, and \emph{responsible}, the next generation L$^3$Ms need to make considerable progress on the following problems that are particularly acute for a computational attorney. 
\begin{itemize}
  \item \emph{Updatable.}
  These models must be kept fresh and current in the relevant legal field if they are to remain reliable for legal reasoning.
  In particular, common law (case law) systems rely heavily on judicial precedents, so outdated or incomplete models may produce wrong legal analysis and results. 
  Taking steps to ensure the timely updating of the models with new information from Westlaw ``Court Wire'' and ``KeyCite Overruling Risk'' etc. can therefore make a considerable difference in their ability to deliver sensible solutions in legal contexts.
  Since retraining the model from scratch to intentionally forget obsolete knowledge and incorporate additional desiderata would be very expensive (costing weeks of time and millions of dollars), we would instead prefer new methods to update the model \emph{post-deployment}, which could yield significant savings over the simplistic strategy of retraining. 
  There have been some early explorations of editing large language models~\cite{mengLocatingEditingFactual2023,mengMassEditingMemoryTransformer2022}. 
  The techniques that have been developed for \emph{stream data mining}, \emph{machine unlearning}, and \emph{life-long learning} (aka \emph{never-ending learning} or \emph{continual learning}) could be helpful. 
  \item \emph{Stable.}
  These models must know their limits and reason within the bounds of existing legal system in the relevant jurisdiction. 
  In other words, we should take measures to prevent the models from ``\emph{hallucinating}'', i.e., inventing seemingly-plausible but non-existent responses by themselves.
  The techniques that have been developed for the quantification of \emph{model uncertainty} aka \emph{epistemic uncertainty} (such as \emph{Bayesian deep learning} and \emph{evidential deep learning}) and \emph{stable learning} for \emph{out-of-distribution} (OOD) generalization based on causality etc. might be useful.
  \item \emph{Provable.}
  These models must be able to prove that their legal opinions or judgments are derived by strictly following the relevant rules as defined in the law~\cite{zhongHowDoesNLP2020}, especially when they are challenged by the other side in court.
  This is a higher requirement than being explainable or being responsible, as the models not only have to explain what logical steps have been taken in their sophisticated legal reasoning process, they also need to identify the concrete applicable legislation to justify the correctness and fairness of each such step.  
  It would not be good enough to just ``get the job done'' or find what training instances or features are more important than others. 
  Probably we can borrow and adapt some techniques from the field of \emph{abductive reasoning} and \emph{neuro-symbolic inference} that have been successful in the mathematical and medical domains. 
  \item \emph{Communicable.}
  These models must communicate effectively with fellow lawyers as well as legal clients to capture the subtle details and nuances of their requirements which may turn out to be crucial in court later. 
  Furthermore, they should be able to learn not only from legal text documents but also directly from human instructions (i.e., be taught by human lawyers).
  ChatGPT has demonstrated the power of \emph{reinforcement learning} from human feedback (RLHF)~\cite{ouyangTrainingLanguageModels2022}, but it is mostly passive rather than active/proactive. 
  In addition to reinforcement learning, \emph{active learning} and \emph{socially situated AI}~\cite{krishnaSociallySituatedArtificial2022} have potential to play a role in this context as well.
  \item \emph{Predictable.}
  These models must be predictable in the sense that they should not just be able to provide valuable legal advice based on sophisticated legal reasoning, but also to anticipate the implications of their outputs to stakeholders and the potential risks or liabilities associated with them in a legal ecosystem probably consisting of both human lawyers and computational attorneys. 
  In order to win cases in court, the models should also be able to imagine the responses from the opponents and negotiate with their lawyers. 
  This would require the combination of language models with strategic thinking, as illustrated by Meta's CICERO~\cite{bakhtinetalHumanLevelPlayGame2022}. 
  Moreover, the models could suggest the timing and strategies for the optimal settlement of a dispute or the best terms of a contract~\cite{conradScenarioAnalyticsAnalyzing2017}, based on their calculations of \emph{Nash equilibria}. 
  The techniques from \emph{algorithmic game theory} and \emph{multi-agent systems} may come in handy.

\end{itemize}

\section{Conclusion}
\label{sec:Conclusion}

In summary, we argue that despite the recent big advances in L$^3$M for the automation of simple legal tasks such as legal text classification, the vision of a computational attorney capable of complex legal reasoning still serves as a ``lode star'' for data mining and machine learning in the legal domain and beyond. 
It is time for researchers to explore how to make legal or general AI models updatable, stable, provable, communicable, and predictable.
The first two objectives of this endeavor seem to be in sight, while the remaining three could take many years before coming to fruition.  


\bibliographystyle{abbrv}
\bibliography{ref_legal-nlp,ref_foundation-models}

\end{document}